# Ranking of classification algorithms in terms of mean-standard deviation using A-TOPSIS

André G. C. Pacheco[a], Renato A. Krohling[a,b]

[a]Graduate Program in Computer Science, PPGI & [b]Production Engineering Department
UFES - Federal University of Espirito Santo
Av. Fernando Ferrari 514 - CEP 29060-270, Vitória, ES, Brazil
[a]pacheco.comp@gmail.com
[b]krohling.renato@gmail.com

**Abstract**  In classification problems when multiples algorithms are applied to different benchmarks a difficult issue arises, i.e., how can we rank the algorithms? In machine learning it is common run the algorithms several times and then a statistic is calculated in terms of means and standard deviations. In order to compare the performance of the algorithms, it is very common to employ statistical tests. However, these tests may also present limitations, since they consider only the means and not the standard deviations of the obtained results. In this paper, we present the so called A-TOPSIS, based on TOPSIS (Technique for Order Preference by Similarity to Ideal Solution), to solve the problem of ranking and comparing classification algorithms in terms of means and standard deviations. We use two case studies to illustrate the A-TOPSIS for ranking classification algorithms and the results show the suitability of A-TOPSIS to rank the algorithms. The presented approach is general and can be applied to compare the performance of stochastic algorithms in machine learning. Finally, to encourage researchers to use the A-TOPSIS for ranking algorithms we also presented in this work an easy-to-use A-TOPSIS web framework.

## 1 Introduction

In machine learning, more precisely in classification problems, it is common applying different algorithms to many benchmarks several times. Normally, the performance of the algorithms are analyzed by means of mean and standard deviation of some known metric, such as the classification accuracy. Next, we need to compare the algorithms and a difficult question arises: how to compare these algorithms effectively? The first answer to this question is to use the statistical tests, i.e., parametric and/or nonparametric. The statistical tests can detect if there are differences between the performances of the algorithms (Derrac et al., 2011; García et al. 2009). One problem is if there are differences, which algorithm is the best one? which is the second better, and which one is the worst? Using nonparametric statistical tests, it is necessary to make pairwise and multiple comparisons among the algorithms. Obviously, the number of tests required increases greatly with the number of algorithms being analyzed. This is problematic, firstly because of the tiresome work of comparing each pair of algorithms; secondly, and more importantly, the probability of making a mistake increases. In addition, these tests may also present limitations, since they consider only the means and not the standard deviations of the obtained results.

Recently, Krohling, Lourenzutti and Campos (2015) presented a new approach to support the selection of the best algorithms by using the Hellinger distance (Hellinger, 1909; Lourenzutti and Krohling, 2014). This approach, called Hellinger-TOPSIS, provides a rank order of the algorithms in a very easy and direct way, by means of the mean and the standard deviation of the performance of the algorithms. However, the Hellinger-TOPSIS presents some shortcomings. Firstly, the mean and the standard deviation of the performance of the algorithms have the same importance. Usually, the mean of the performance is more important than the standard deviation and in the Hellinger-TOPSIS we can not control the influence of these two parameters. Second, if any algorithm in the group is deterministic, i.e, the results obtained are described just by the means and we need to compare it with others stochastic ones, the Hellinger-TOPSIS cannot handle with such a case, because in the algorithm the standard deviation must be different of zero.

In our previous work, we proposed the A-TOPSIS (Krohling and Pacheco, 2015), a new approach that provides a rank order of the evolutionary algorithms in cases where the performance of the algorithms are expressed in terms of means and standard deviations. In this work, we extend our previous approach by providing an in-depth investigation for two case studies for classification problems and developing an easy-to-use web framework for A-TOPSIS. The remainder of this paper is organized as follows: Section 2 describes the TOPSIS. In Section 3, we present the approach based on TOPSIS to deal with data matrix consisting of performance of algorithms in terms of means and standard deviations and we briefly describe the framework. In Section 4, we present simulation results for two case studies involving the classification task in order to illustrate the suitability of the presented approach. In Section 5, conclusions and directions for future work are given.

## 2 TOPSIS - Technique for Order Preference by Similarity to Ideal Solution

The Technique for Order Preference by Similarity to Ideal Solution (TOPSIS) developed by Hwang & Yoon (1981) is a technique to evaluate the performance of alternatives through the similarity with the ideal solution. According to this technique, the best alternative would be one that is closest to the positive-ideal solution and farthest from the negative-ideal solution. The positive-ideal solution is the one that maximizes the benefit criteria and minimizes the cost criteria. The negative-ideal solution maximizes the cost criteria and minimizes the benefit criteria. In summary, the positive-ideal solution is composed of all the best values attainable for the criteria, and the negative-ideal solution consists of all the worst values attainable for the criteria. The interested reader shall refer to Behzadian et al. (2012) for a broad survey about TOPSIS.

Let us consider the decision matrix $A$, which consists of a*lternatives* and *criteria*, described by:

$$A = \begin{matrix} & \begin{matrix} C_1 & ... & C_n \end{matrix} \\ \begin{matrix} A_1 \\ ... \\ A_m \end{matrix} & \begin{pmatrix} x_{11} & \cdots & x_{1n} \\ \vdots & \ddots & \vdots \\ x_{m1} & \cdots & x_{mn} \end{pmatrix} \end{matrix} \quad (1)$$

where $A_1, A_2, \cdots, A_m$ are viable alternatives, and $C_1, C_2, \cdots, C_n$ are criteria, $x_{ij}$ indicates the rating of the alternative $A_i$ with respect to criterion $C_j$. The weight vector $W = (w_1, w_2, ..., w_n)$ is composed of the individual weights $w_j (j = 1, ..., n)$, for each criterion $C_j$ and satisfies $\sum_{j=1}^{n} w_j = 1$. In general, the criteria can be classified into two types: *benefit* and *cost*. The *benefit* criterion means that a higher value is better, while for the *cost* criterion the opposite is valid. The data of the decision matrix $A$ come from different sources, so it is necessary to normalize it in order to transform it into a dimensionless matrix, which allows the comparison of the various criteria. In this work, we use the normalized decision matrix $R = \left[ r_{ij} \right]_{mxn}$ with $i = 1, ..., m$, and $j = 1, ..., n$. The normalized value $r_{ij}$ is calculated as:

$$r_{ij} = \frac{x_{ij}}{\sqrt{\sum_{i=1}^{m} x_{ij}^2}}, \text{ with } i = 1, ..., m; j = 1, ..., n. \quad (2)$$

or

$$r_{ij} = \frac{x_{ij}}{x_{i\max}}, \text{ with } i = 1, ..., m; j = 1, ..., n. \quad (3)$$

The normalized decision matrix $R$ represents the relative rating of the alternatives. After normalization, one calculates the weighted normalized decision matrix $P = \left[ p_{ij} \right]_{mxn}$ with $i = 1, ..., m$, and $j = 1, ..., n$ by multiplying the normalized decision matrix by its associated weights. The weighted normalized value $p_{ij}$ is calculated as:

$$p_{ij} = w_j \cdot r_{ij} \text{ with } i = 1,...,m, \text{ and } j = 1,...,n. \tag{4}$$

The TOPSIS is described in the following steps (Hwang and Yoon, 1981; Krohling and Campanharo, 2011):

**Step 1:** Identify the positive ideal solutions $A^+$ (benefits) and negative ideal solutions $A^-$ (costs) as follows:

$$A^+ = (p_1^+, p_2^+, ..., p_n^+) \tag{5}$$

$$A^- = (p_1^-, p_2^-, ..., p_n^-) \tag{6}$$

where

$$p_j^+ = \left( \max_i p_{ij}, j \in J_1; \min_i p_{ij}, j \in J_2 \right)$$

$$p_j^- = \left( \min_i p_{ij}, j \in J_1; \max_i p_{ij}, j \in J_2 \right)$$

$J_1$ and $J_2$ represent the criteria *benefit* and *cost*, respectively.

**Step 2:** Calculate the Euclidean distances from the positive ideal solution $A^+$ (benefits) and the negative ideal solution $A^-$ (costs) of each alternative $A_i$, respectively as follows:

$$d_i^+ = \sqrt{\sum_{j=1}^n (d_{ij}^+)^2} \tag{7}$$

$$d_i^- = \sqrt{\sum_{j=1}^n (d_{ij}^-)^2} \tag{8}$$

where

$d_{ij}^+ = p_j^+ - p_{ij}$, with $i = 1,...,m$.

$d_{ij}^- = p_j^- - p_{ij}$, with $i = 1,...,m$.

**Step 3:** Calculate the relative closeness coefficients $\xi_i$ for each alternative $A_i$ with respect to the positive ideal solution as given by:

$$\xi_i = \frac{d_i^-}{d_i^+ + d_i^-} \tag{9}$$

**Step 4:** Rank the alternatives according to the relative closeness. The best alternatives are those that have higher value $\xi_i$ and therefore should be chosen.

Next, we describe the A-TOPSIS approach involving two matrices.

## 3 A-TOPSIS - An approach based on TOPSIS for ranking algorithms

The decision matrix consisting of a*lternatives* and *criteria* is described by

$$\tilde{D} = \begin{pmatrix} x_{11} & \cdots & x_{1n} \\ \vdots & \ddots & \vdots \\ x_{m1} & \cdots & x_{mn} \end{pmatrix} = \begin{pmatrix} (\mu_{11}, \sigma_{11}) & \cdots & (\mu_{1n}, \sigma_{1n}) \\ \vdots & \ddots & \vdots \\ (\mu_{m1}, \sigma_{m1}) & \cdots & (\mu_{mn}, \sigma_{mn}) \end{pmatrix}$$

where $A_1, A_2, \cdots, A_m$ are alternatives, $C_1, C_2, ..., C_n$ are criteria, $x_{ij}$ indicates the rating of the alternative $A_i$ with respect to criterion $C_j$ described in terms of its mean and standard deviations $\mu_{ij}$, $\sigma_{ij}$, respectively. In the context of comparison of algorithms, the alternatives consists of the algorithms and the criteria are the benchmark problems.

One could interpret this problem as consisting of two decision matrices given by $D = \{M_\mu, M_\sigma\}$.

$$M_\mu = \begin{pmatrix} \mu_{11} & \cdots & \mu_{1n} \\ \vdots & \ddots & \vdots \\ \mu_{m1} & \cdots & \mu_{mn} \end{pmatrix} \quad M_\sigma = \begin{pmatrix} \sigma_{11} & \cdots & \sigma_{1n} \\ \vdots & \ddots & \vdots \\ \sigma_{m1} & \cdots & \sigma_{mn} \end{pmatrix}$$

In this context, we develop a new framework combining the TOPSIS for ranking algorithms in terms of mean and standard deviations as illustrated in Fig. 1.

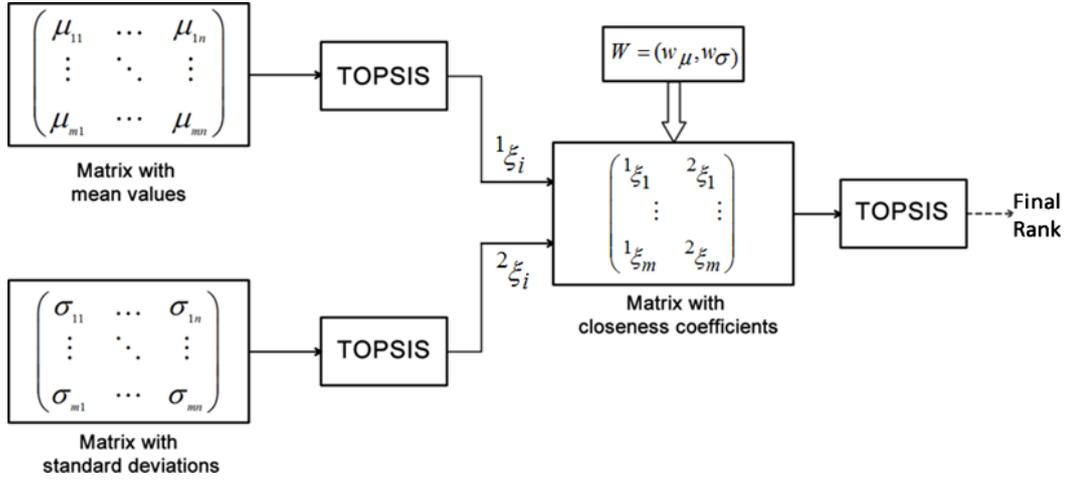

Figure 1: *Illustration of TOPSIS approach for ranking algorithms in terms of mean and standard deviations.*

### 3.1 A-TOPSIS algorithm

The steps of the algorithm are described as follows:

**Step 1:** Normalize the matrices $M_\mu$ and $M_\sigma$.

**Step 2:** Identify the positive ideal solutions $A^+$ (benefits) and negative ideal solutions $A^-$ (costs) for each matrix as follows:

$$A^+ = (p_1^+, p_2^+, ..., p_n^+) \tag{10}$$

$$A^- = (p_1^-, p_2^-, ..., p_n^-) \tag{11}$$

where

$$p_j^+ = \left( \max_i p_{ij}, j \in J_1; \min_i p_{ij}, j \in J_2 \right)$$

$$p_j^- = \left( \min_i p_{ij}, j \in J_1; \max_i p_{ij}, j \in J_2 \right)$$

$J_1$ and $J_2$ represent the criteria *benefit* and *cost*, respectively.

**Step 3:** Calculate the Euclidean distances from the positive ideal solution $A^+$ (benefits) and the negative ideal solution $A^-$ of each alternative $A_i$, respectively as follows:

$$d_i^+ = \sqrt{\sum_{j=1}^{n}(p_j^+ - p_{ij},)^2} \quad \text{with } i=1,...,m. \tag{12}$$

$$d_i^- = \sqrt{\sum_{j=1}^{n}(p_j^- - p_{ij})^2} \quad \text{with } i=1,...,m. \tag{13}$$

**Step 4:** Calculate the relative closeness coefficients for each alternative $\xi_i$ with respect to positive ideal solution as:

$$\xi_i = \frac{d_i^-}{d_i^+ + d_i^-} \quad \text{with } i=1,...,m. \tag{14}$$

**Step 5:** After calculating the vector $\xi_i$ for both decision matrices, we obtain a data matrix that is made up of the two vectors of the relative closeness coefficients, as given by:

$$C = \begin{pmatrix} {}^1\xi_1 & {}^2\xi_1 \\ \vdots & \vdots \\ {}^1\xi_m & {}^2\xi_m \end{pmatrix} \tag{15}$$

In this case, to each of the vectors, it is assigned a weight $W = (w_1, w_2) = (w_\mu, w_\sigma)$, where $w_\mu$ and $w_\sigma$ represent the weight assigned to the criteria means, and standard deviations, respectively, which satisfies $w_\mu + w_\sigma = 1$. One can now obtain the weighted relative-closeness coefficients matrix by introducing the importance weights to each one of the relative-closeness coefficient vector, as given by:

$$C = \begin{pmatrix} w_1\,{}^1\xi_1 & w_2\,{}^2\xi_1 \\ \vdots & \vdots \\ w_1\,{}^1\xi_m & w_2\,{}^2\xi_m \end{pmatrix} \tag{16}$$

From this stage on, the method continues by applying the standard TOPSIS to the resulting matrix in order to identify the global ranking.

**Step 6:** Identify the global positive ideal solution $A_G^+$ and the global negative ideal solution $A_G^-$, respectively, as follows:

$$A_G^+ = (p_{G1}^+, p_{G2}^+) = \left( \max_i {}^l\xi_i,\ l \in J_1; \min_i {}^l\xi_i\ l \in J_2 \right). \tag{17}$$

$$A_G^- = (p_{G1}^-, p_{G2}^-) = \left( \min_i {}^l\xi_i,\ l \in J_1; \max_i {}^l\xi_i\ l \in J_2 \right). \tag{18}$$

where $J_1$ and $J_2$ represent the criteria *benefit* and *cost*, respectively.

**Step 7:** Calculate to each alternative $A_i$ the distances from the global positive ideal solution $A_G^+$ and from the global negative ideal solution $A_G^-$, respectively, as follows:

$$d_{Gi}^+ = \sqrt{\sum_{l=1}^{2}({}^l\xi_i - p_{G_l}^+)^2} \quad \text{with } i=1,....,m. \tag{19}$$

$$d_{Gi}^- = \sqrt{\sum_{l=1}^{2}({}^l\xi_i - p_{G_l}^-)^2} \quad \text{with } i=1,....,m. \tag{20}$$

**Step 8:** Calculate the global relative-closeness coefficients $\xi_{Gi}$ for each alternative $A_i$ with respect to global positive ideal solution $A_G^+$ as:

$$\xi_{Gi} = \frac{d_{Gi}^-}{d_{Gi}^- + d_{Gi}^+} \quad (21)$$

**Step 9:** Rank the alternatives according to the relative closeness coefficients. The best alternatives are those that have higher value $\xi_{Gi}$ and therefore should be chosen.

### 3.2 A-TOPSIS web framework

In order to encourage researchers and practitioners in different areas of knowledge to use the A-TOPSIS for ranking algorithms, we provide an easy-to-use web framework. As shown in figure 2, to use this framework the user needs to set the matrices $M_\mu$ and $M_\sigma$ as .csv files and the value of the weights for each one. Thereby, the framework provides the graph bar rank and the values of the closeness coefficients.

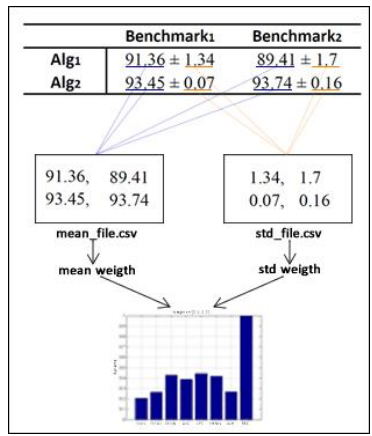

*Figure 2: The A-TOPSIS framework*

The A-TOPSIS framework can be easily used by accessing the web address http://www.inf.ufes.br/~agcpacheco/alg-ranking/.

## 3 Simulation Results

In this section, we present two case studies involving classification problems. In order to compare our results, we also apply the Hellinger-TOPSIS for each case and we used the non-parametric Friedman test followed by Wilcoxon test as a pos hoc, both with $p_{value} = 0.05$ (Derrac et al., 2011). As the Hellinger-TOPSIS is not able to handle with standard deviation equal to zero, we set a very small value as standard deviation in cases where this occur.

### 3.1 Case study I

In this case study, we have an ensemble of classifiers, containing four classifiers: feedforward neural network (FNN), extreme learning machine (ELM), discriminative restricted Boltzmann machine (DRBM) and K-nearest neighbors (KNN). In addition, we have three aggregation methodologies: average of the supports (AVG), majority voting (MV) and Choquet integral (CHO). So, the goal is to obtain a rank of the classifiers according to 12 benchmarks. In this case, the alternatives are the classifiers and the criteria are the benchmarks. In table 1 is shown the performance of the classifiers, in terms of mean and standard deviation of the classification accuracy. Obviously, to apply A-TOPSIS for this case the criterion is set as benefit, i.e., the higher the value is, the better.

As we can see in table 1, the KNN algorithm does not have standard deviation, it was used with just one value of K. Because of this, we divided this study in two parts: first, we remove the KNN and consider the remaining classifiers. Secondly, we consider the standard deviation of the KNN as zero. Next, we carry out a sensitivity study by varying the weights for mean and standard deviation, respectively, and present the results for each part of the study.

**3.1.1 Case study I – Part I**

Removing the KNN of the decision matrix, in table 2 we present the results of the rank by varying the weights.

| Weight variation [mean, std] | Ranking |
|---|---|
| [0.5, 0.5] | CHO ≻ MV ≻ AVG ≻ ELM ≻ DRBM ≻ FNN |
| [0.6, 0.4] | CHO ≻ MV ≻ AVG ≻ FNN ≻ ELM ≻ DRBM |
| [0.7, 0.3] | CHO ≻ MV ≻ AVG ≻ FNN ≻ ELM ≻ DRBM |
| [0.8, 0.2] | CHO ≻ MV ≻ AVG ≻ FNN ≻ ELM ≻ DRBM |
| [0.1, 0.9] | CHO ≻ MV ≻ AVG ≻ FNN ≻ ELM ≻ DRBM |
| [1, 0] | CHO ≻ MV ≻ AVG ≻ FNN ≻ ELM ≻ DRBM |

*Table 2: Rank by varying the values of the weight – Case study I, Part I*

As we can see, the first, second and third place in the rank do not change regardless the weight. In fact, the only change in the rank occurs when the values of the weights become [0.6, 0.4]. In this case, the FNN rises to the fourth place, the ELM goes down to the second place and the DRBM goes to the last place. From the values of the weight [0.6, 0.4] to [1, 0] the rank keeps the same. In figure 3 is illustrated the raking in bar graph for each value of weights for this part of the study.

| Classifiers | Benchmarks | | | | | | | | | | | |
|---|---|---|---|---|---|---|---|---|---|---|---|---|
| | **Susy** | **Higgs** | **Covtype** | **DNA** | **Isolet** | **Cancer** | **Cred. Aus** | **Diabetic** | **Iris** | **Spam** | **Statlog** | **Wine** |
| **FNN** | 78,14 ± 0,65 | 63,21 ± 1,19 | 75,22 ± 1,09 | 91,36 ± 1,34 | 89,41 ± 1,7 | 94,87 ± 0,62 | 83,05 ± 1,07 | 71,02 ± 2,31 | 95,62 ± 1,36 | 93,61 ± 0,8 | 99,59 ± 0,06 | 95,91 ± 2,16 |
| **DRBM** | 76,39 ± 0,32 | 63,4 ± 0,3 | 66,25 ± 0,17 | 93,45 ± 0,07 | 93,74 ± 0,16 | 95,39 ± 0,28 | 82,88 ± 0,8 | 60,48 ± 1,69 | 89,4 ± 0,95 | 90,8 ± 0,56 | 92,08 ± 0,04 | 95,15 ± 1,28 |
| **ELM** | 79,39 ± 0,29 | 63,99 ± 0,09 | 76,01 ± 0,11 | 90,59 ± 0,75 | 86,81 ± 0,6 | 95,07 ± 0,5 | 83,38 ± 0,81 | 72,37 ± 1,09 | 94,81 ± 1,96 | 89,43 ± 0,57 | 98,3 ± 0,08 | 91,13 ± 2,8 |
| **KNN** | 70,88 ± 0 | 59,84 ± 0 | 75,81 ± 0 | 85,98 ± 0 | 88,24 ± 0 | 95,23 ± 0 | 67,63 ± 0 | 61,73 ± 0 | 95,55 ± 0 | 72,75 ± 0 | 98,73 ± 0 | 67,92 ± 0 |
| **AVG** | 78,38 ± 0,59 | 64,01 ± 0,96 | 71,84 ± 0,42 | 92,18 ± 0,92 | 93,73 ± 0,26 | 95,38 ± 0,35 | 83,14 ± 0,55 | 68,84 ± 3,03 | 95,11 ± 1,47 | 93,4 ± 0,61 | 98,73 ± 0,01 | 96,16 ± 1,44 |
| **MV** | 78,14 ± 0,54 | 63,75 ± 0,64 | 75,75 ± 0,24 | 92,64 ± 0,57 | 93,93 ± 0,31 | 94,93 ± 0,42 | 83,15 ± 0,98 | 70,05 ± 2,24 | 95,62 ± 1,42 | 92,02 ± 0,37 | 99,32 ± 0,04 | 95,22 ± 1,76 |
| **CHO** | 78,58 ± 0,51 | 64,7 ± 0,72 | 76,85 ± 0,29 | 93,69 ± 0,38 | 93,75 ± 0,16 | 95,57 ± 0,35 | 83,52 ± 0,85 | 71,42 ± 0,97 | 95,7 ± 1,29 | 93,78 ± 0,65 | 98,73 ± 0,01 | 97,18 ± 1,34 |

*Table 1: The classifiers performance for each benchmark – Case study I*

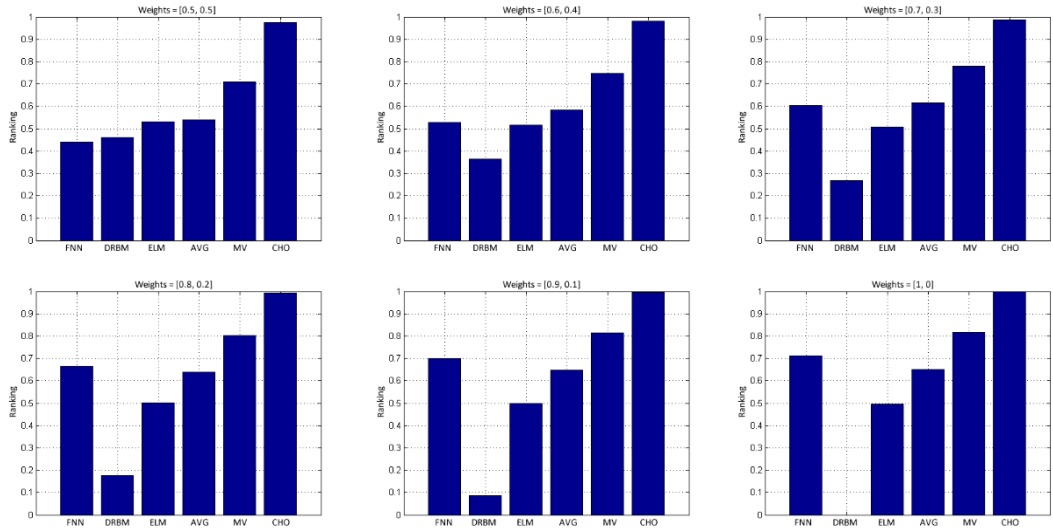

*Figure 3: Rank in bar graph for each values of the weights – Case study I, Part I*

We compare the results obtained by A-TOPSIS with the Hellinger - TOPSIS. In this case, the rank become stable when the values of the weights are [0.6, 0.4]. Therefore, we decide to choose these weights. In table 3 is presented the rank for each methodology, which is also depicted in figure 4 by means of bar graph. According to the results of the table 3, both methods obtained the same rank.

| Method | Ranking |
| --- | --- |
| A-TOPSIS | CHO $\succ$ MV $\succ$ AVG $\succ$ FNN $\succ$ ELM $\succ$ DRBM |
| Hellinger -TOPSIS | CHO $\succ$ MV $\succ$ AVG $\succ$ FNN $\succ$ ELM $\succ$ DRBM |

*Table 3: Rank comparison between A-TOPSIS and H-TOPSIS – Part I*

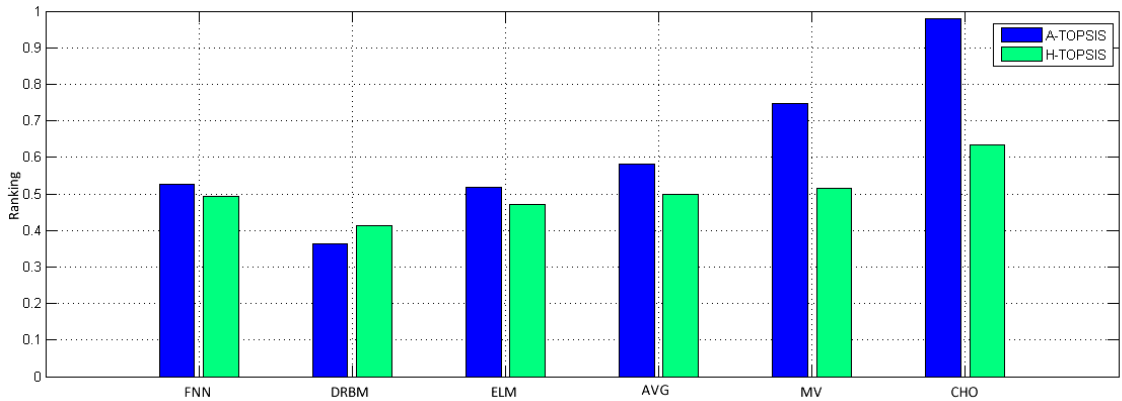

*Figure 4: Rank in bar graph for A-TOPSIS and H-TOPSIS – Case study I, Part I*

The Friedman test for this part of the case study I provides $p_{value} = 0.00005$ leading to reject $H_0$. Then, we performe pairwise comparisons by means of the Wilcoxon test, presented in table 4. According to the results of the Wilcoxon test, the CHO classifier is significantly different comparing to the other ones. Furthermore, the DRBM classifier is significantly different comparing to AVG, MV and CHO. The results of the statistical tests indicate that in this case, the CHO classifier is the best and the DRBM is the worst one. This finding is consistent with the results obtained by A-TOPSIS. Nonetheless, this statistical test cannot provide a rank with all the classifiers as A-TOPSIS does.

| Pairwise | p | Pairwise | p |
|---|---|---|---|
| FNN - CHO | 0.009277 | ELM - CHO | 0.042480 |
| DRBM - AVG | 0.015137 | AVG - CHO | 0.000977 |
| DRBM - MV | 0.026855 | MV - CHO | 0.009277 |
| DRBM - CHO | 0.000488 | - | - |

*Table 4: Wilcoxon test pairwise comparison with p less than 0.05 – Case study I, part I*

### 3.1.2 Case study I - Part II

Considering the KNN in table 1, the results of the rank by varying the weights are presented in table 5.

| Weight variation [mean, std] | Ranking |
|---|---|
| [0.5, 0.5] | CHO ≻ MV ≻ KNN ≻ DRBM ≻ AVG > ELM ≻ FNN |
| [0.6, 0.4] | CHO ≻ MV ≻ AVG ≻ ELM ≻ FNN ≻ DRBM ≻ KNN |
| [0.7, 0.3] | CHO ≻ MV ≻ AVG ≻ FNN ≻ ELM ≻ DRBM ≻ KNN |
| [0.8, 0.2] | CHO ≻ MV ≻ AVG ≻ FNN ≻ ELM ≻ DRBM ≻ KNN |
| [0.1, 0.9] | CHO ≻ MV ≻ AVG ≻ FNN ≻ ELM ≻ DRBM ≻ KNN |
| [1, 0] | CHO ≻ MV ≻ AVG ≻ FNN ≻ ELM ≻ DRBM ≻ KNN |

*Table 5: Rank by varying the values of the weights – Case study I, Part II*

As we can notice in table 5, for all weights the first and the second place in the rank do not change. When the weights of the mean and the standard deviation are equals, the KNN reaches the third place in the rank. However, when the values of the weights become [0.6, 0.4], only 10% of variation, the KNN goes to the last place. Moreover, for these values of the weights, the ELM rises to the fourth place. Nevertheless, when the values turn to [0.7, 0.3], the ELM and the FNN change its positions. From the values of the weights [0.7, 0.3] to [1, 0], the rank does not change. In figure 5 is illustrated the rank in bar graph for each value of weights for this part of the study.

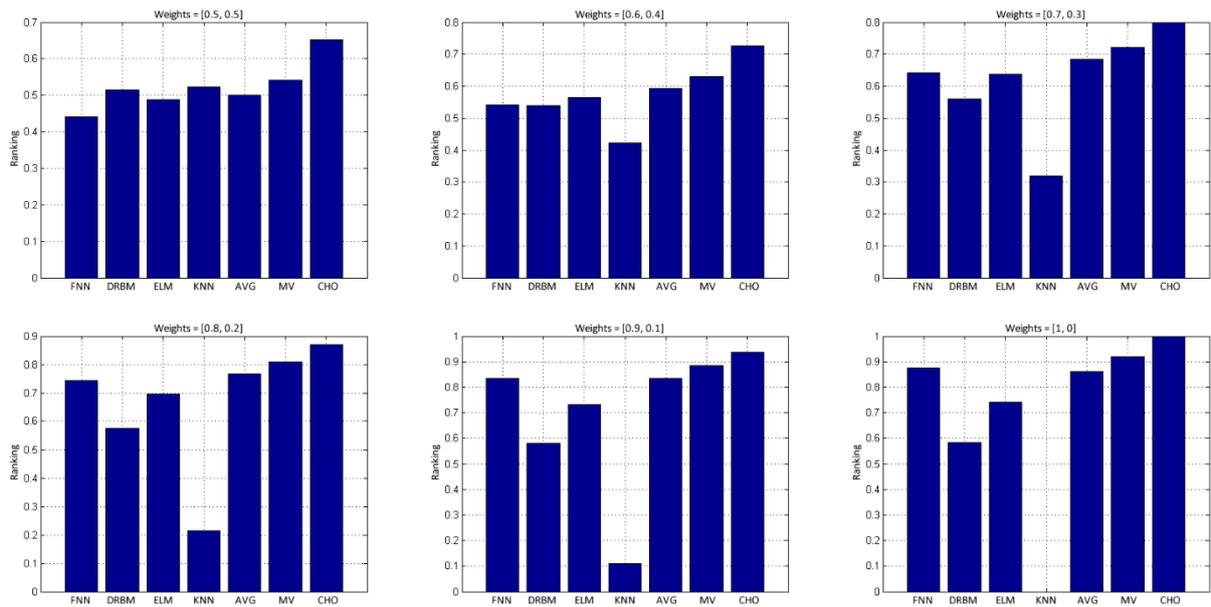

*Figure 5: Rank in bar graph for each values of the weight – Case study I, Part II*

Finally, we compare the results obtained by A-TOPSIS with the Hellinger TOPSIS. We decide to choose the values of the weights as [0.7, 0.3]. As shown in table 5, for these values of the weights, the rank become stable and does not change when we vary the weights. The rank obtained for A-TOPSIS and Hellinger-TOPSIS is presented in table 6 and illustrated in figure 6.

| Method | Ranking |
| --- | --- |
| A-TOPSIS | CHO ≻ MV ≻ AVG ≻ FNN ≻ ELM ≻ DRBM ≻ KNN |
| Hellinger - TOPSIS | CHO ≻ ELM ≻ MV ≻ FNN ≻ AVG ≻ DRBM ≻ KNN |

*Table 6: Rank comparison between A-TOPSIS and H-TOPSIS – Case study I, Part II*

In table 6, we can easily check that the ranking of the alternatives CHO, FNN, DRBM and KNN are the same in both methods. Thereby, the ranking of the best and worst alternatives are kept. On the other hand, the ranking of the alternatives MV, AVG and ELM have changed position.

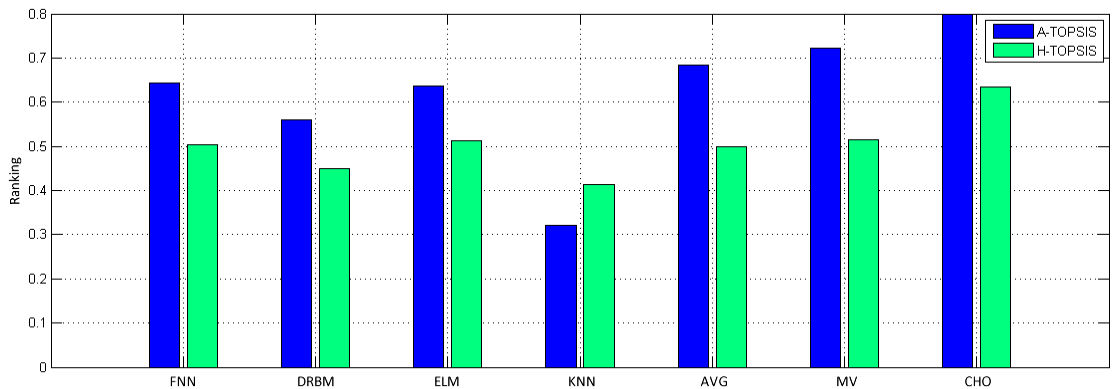

*Figure 6: Rank in bar graph for A-TOPSIS and H-TOPSIS – Case study I, Part II*

Similarly to the previous part, we performed the Friedman test, which obtained $p_{value} = 0.00007$, leading to reject $H_0$. Next, we applied the Wilcoxon test and the results are shown in table 7. As in the part I, the CHO classifier is significantly different comparing to the other ones, according to Wilcoxon test. In addition, the KNN is also significantly different from the others, except DRBM. Finally, the DRBM is significantly different comparing to AVG, MV, CHO. The results of the statistical tests indicate that the best classifier is CHO and the worst are the KNN and DRBM. Also in this case, this finding is consistent with the rank obtained by A-TOPSIS.

| Pairwise | p | Pairwise | p |
|---|---|---|---|
| FNN - KNN | 0.004883 | ELM - CHO | 0.042480 |
| FNN - CHO | 0.009277 | KNN - AVG | 0.009766 |
| DRBM - AVG | 0.015137 | KNN - MV | 0.003418 |
| DRBM - MV | 0.026855 | KNN - CHO | 0.000977 |
| DRBM - CHO | 0.000488 | AVG - CHO | 0.000977 |
| ELM - KNN | 0.042480 | MV - CHO | 0.009277 |

*Table 7: Wilcoxon test pairwise comparison with p less than 0.05 – Case study I, part II*

### 3.2 Case study II

This case study presented by Wen et al. (2013) consists in a classification problem with 8 classifiers performed to 10 benchmarks as shown in table 8. Similar to the case study 1, our goal is to find the rank of the classifiers according to the mean and the standard deviation of the classifiers performance. It is worth mentioning that in this case study the authors used the error rate as accuracy. The A-TOPSIS can easily handle with this just changing the criterion from benefit to cost, i.e., the smaller the value is, the better.

Firstly, we carry out a sensitivity study by varying the weights for mean and standard deviation, respectively, and the rank of the classifiers for each weight is shown in table 9 and in figure 7 by means of bar graph.

| Weight variation [mean, std] | Ranking |
|---|---|
| [0.5, 0.5] | REC ≻ LPC ≻ EKNN ≻ HKNN ≻ LNC ≻ ALH ≻ FKNN ≻ KNN |
| [0.6, 0.4] | REC ≻ HKNN ≻ LNC ≻ LPC ≻ EKNN ≻ ALH ≻ FKNN ≻ KNN |
| [0.7, 0.3] | REC ≻ HKNN ≻ LNC ≻ LPC ≻ EKNN ≻ ALH ≻ FKNN ≻ KNN |
| [0.8, 0.2] | REC ≻ HKNN ≻ LNC ≻ LPC ≻ EKNN ≻ ALH ≻ FKNN ≻ KNN |
| [0.9, 0.1] | REC ≻ HKNN ≻ LNC ≻ LPC ≻ EKNN ≻ ALH ≻ FKNN ≻ KNN |
| [1, 0] | REC ≻ HKNN ≻ LNC ≻ LPC ≻ ALH ≻ EKNN ≻ FKNN ≻ KNN |

*Table 9: Rank by varying the values of the weights – Case study II*

As we can notice in table 9 and in figure 7, the first and the last place are the same for all weights. When the weights are varied from [0.5, 0.5] to [0.6, 0.4] the positions of the classifiers HKNN, LNC, LPC and EKNN are changed. For the values of the weights from [0.6, 0.4] to [0.9, 0.1] the rank does not change. Finally, when the weights become [1, 0] the classifiers EKNN and ALH switch their positions.

| Classifiers | Benchmarks | | | | | | | | | |
|---|---|---|---|---|---|---|---|---|---|---|
| | **B1** | **B2** | **B3** | **B4** | **B5** | **B6** | **B7** | **B8** | **B9** | **B10** |
| **KNN** | 3.49 ± 0.49 | 3.35 ± 0.39 | 25.83 ± 0.70 | 19.42 ± 0.79 | 30.82 ± 1.40 | 14.10 ± 1.60 | 4.40 ± 0.60 | 3.94 ± 0.37 | 18.96 ± 1.71 | 29.84 ± 0.55 |
| **FKNN** | 3.49 ± 0.24 | 3.13 ± 0.30 | 26.28 ± 0.61 | 15.96 ± 0.68 | 30.45 ± 1.82 | 14.10 ± 1.04 | 4.53 ± 0.56 | 2.40 ± 0.36 | 19.04 ± 0.93 | 30.50 ± 1.03 |
| **EKNN** | 2.26 ± 0.56 | 2.96 ± 0.57 | 25.76 ± 1.01 | 10.77 ± 1.25 | 30.71 ± 0.51 | 14.10 ± 1.24 | 5.07 ± 0.37 | 3.90 ± 0.32 | 19.19 ± 1.21 | 31.16 ± 1.40 |
| **LMC** | 2.70 ± 0.45 | 2.35 ± 0.47 | 24.85 ± 0.86 | 11.11 ± 1.05 | 31.59 ± 1.84 | 11.90 ± 1.47 | 4.27 ± 0.76 | 2.00 ± 0.24 | 19.41 ± 1.90 | 26.84 ± 1.08 |
| **LPC** | 2.60 ± 0.49 | 2.68 ± 0.58 | 25.03 ± 0.68 | 12.36 ± 0.59 | 32.0 ± 1.83 | 11.90 ± 1.21 | 4.00 ± 0.82 | 2.10 ± 0.31 | 19.04 ± 1.0 | 27.10 ± 0.40 |
| **HKNN** | 2.04 ± 0.84 | 2.57 ± 0.73 | 25.31 ± 0.83 | 11.44 ± 0.55 | 29.63 ± 2.34 | 11.81 ± 1.66 | 4.00 ± 0.82 | 1.98 ± 0.49 | 20.44 ± 1.29 | 24.64 ± 1.28 |
| **ALH** | 2.48 ± 0.51 | 3.07 ± 0.44 | 29.92 ± 0.88 | 11.62 ± 0.79 | 31.37 ± 2.36 | 10.86 ± 1.28 | 4.67 ± 0.94 | 2.00 ± 0.32 | 22.30 ± 1.27 | 26.72 ± 1.20 |
| **REC** | 1.58 ± 0.46 | 2.13 ± 0.32 | 24.35 ± 0.75 | 6.10 ± 0.43 | 28.48 ± 1.20 | 10.76 ± 0.93 | 4.00 ± 0.00 | 0.79 ± 0.13 | 18.52 ± 0.74 | 24.40 ± 0.90 |

*Table 8: The classifiers performance for each benchmark - Case study 2*

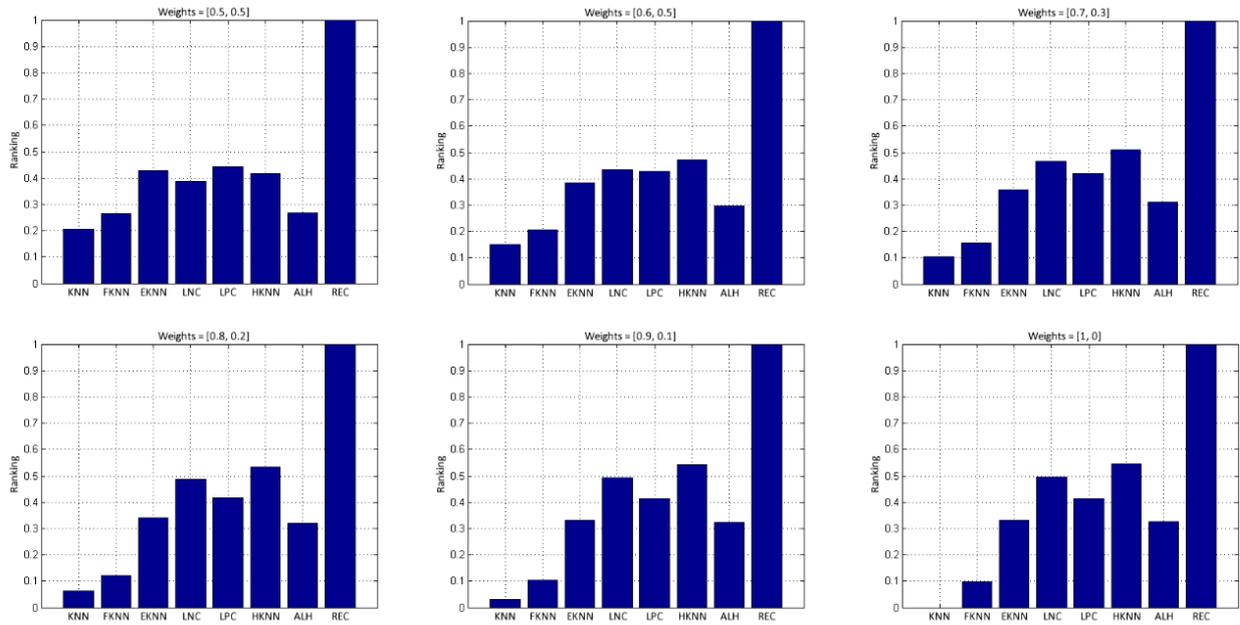

*Figure 7: Rank in bar graph for each values of the weight – Case study II*

Again, we compare the results obtained by A-TOPSIS with the Hellinger - TOPSIS. We decided to choose the values of the weights as [0.7, 0.3]. As shown in table 9, for these values, the rank become stable and does not change when we vary the weights. The rank obtained for A-TOPSIS and Hellinger-TOPSIS is presented in table 10 and illustrated in figure 8 by means of bar graph.

| Method | Ranking |
| --- | --- |
| A-TOPSIS | REC $\succ$ HKNN $\succ$ LNC $\succ$ LPC $\succ$ EKNN $\succ$ ALH $\succ$ FKNN $\succ$ KNN |
| Hellinger - TOPSIS | REC $\succ$ LNC $\approx$ HKNN $\succ$ LPC $\succ$ ALH $\succ$ FKNN $\succ$ EKNN $\succ$ KNN |

*Table 10: Rank comparison between A-TOPSIS and H-TOPSIS – Case study II*

In table 10, we can notice that there are some differences in the ranks. For both ranks, the first and the last places are kept. In the Hellinger-TOPSIS, the ranking for the classifiers LNC and HKNN are too close, because of this, they are tied in second place. Furthermore, as we can see in figure 8, even though we can distinguish the ranking for the classifiers FKNN, EKNN, LPC and ALH, the values of the closeness coefficients are too close. This issue does not occur in A-TOPSIS.

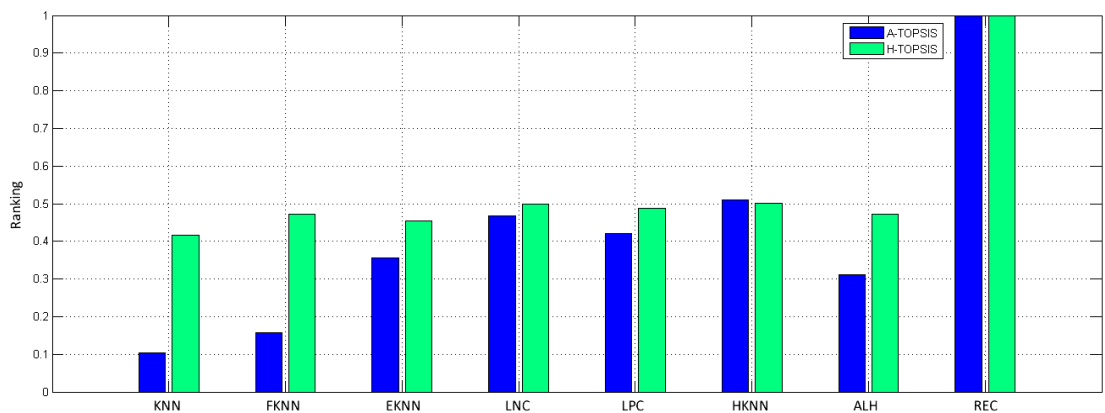

*Figure 8: Rank in bar graph for A-TOPSIS and H-TOPSIS – Case study II*

To the case study II we also performed the Friedman test that provides $p_{value} = 0.00001$, leading to reject $H_0$ one more time. The pos hoc obtained by Wilcoxon test is shown in table 11. In this case study, the classifier REC is significantly different comparing to the other ones. Moreover, the KNN classifier is significantly different from LMC, LPC, HKNN and REC, and the FKNN from LMC, HKNN, REC and ALH. The statistical tests point that the best classifier in the group is REC and the worst are KNN and EKNN. In this case study, the results obtained by the statistical test and the A-TOPSIS are consistent.

| **Pairwise** | *p* | **Pairwise** | *p* |
|---|---|---|---|
| KNN - LMC | 0.019531 | EKNN - REC | 0.001953 |
| KNN - LPC | 0.037109 | LMC - REC | 0.001953 |
| KNN - HKNN | 0.027344 | LPC - REC | 0.003906 |
| KNN - REC | 0.001953 | HKNN - ALH | 0.027344 |
| FKNN - LMC | 0.048828 | HKNN - REC | 0.003906 |
| FKNN - HKNN | 0.027344 | ALH - REC | 0.001953 |
| FKNN - REC | 0.001953 | - | - |

*Table 11: Wilcoxon test pairwise comparison with p less than 0.05 – Case study II*

## 4 Concluding remarks

In this work, we present an application of the A-TOPSIS algorithm to compare performance of classification algorithms by means of the mean and the standard deviation. In order to illustrate the method, two case studies involving classification problems is presented. The rank provided by A-TOPSIS is compared with Hellinger-TOPSIS and we carried out the nonparametric statistical tests of Friedman and Wilcoxon in order to analyze the rank order. The results for the case studies show the effectiveness of the method. Despite the case studies are only for classification problems, the presented approach is general and can be applied to compare the performance of stochastic algorithms in machine learning.

In terms of computational burden, the A-TOPSIS consists of a very simple computation procedure. It is worth to note that, the TOPSIS is a well-established and reliable methodology, which guarantee the A-TOPSIS effectiveness. Finally, in order to encourage researchers and practitioners in different areas of knowledge, especially in machine learning, to use the A-TOPSIS, we provided a web framework to rank algorithms in an easy way.

**Acknowledgments**. A.G.C. Pacheco would like to thank the financial support of the Brazilian agency CAPES and R.A. Krohling thanks the financial support of the Brazilian agency CNPq under grant nr. 309161/2015-0.